\documentclass[conference]{IEEEtran}
\IEEEoverridecommandlockouts
\usepackage{cite}
\usepackage{amsmath,amssymb,amsfonts}
\usepackage{algorithm} 
\usepackage{algorithmic}
\usepackage{graphicx}
\usepackage{mathrsfs}
\usepackage{textcomp}
\usepackage{xcolor}
\usepackage{float}  
\usepackage{caption}
\usepackage{subcaption}
\usepackage[caption=false]{subfig}
\usepackage{adjustbox} 
\usepackage[hyphens]{url}
\usepackage{fancyhdr}
\def\BibTeX{{\rm B\kern-.05em{\sc i\kern-.025em b}\kern-.08em
    T\kern-.1667em\lower.7ex\hbox{E}\kern-.125emX}}

\begin{document}

\title{LLM-OptiRA: LLM-Driven Optimization of Resource Allocation for Non-Convex Problems in Wireless Communications\\
}

	\author{\IEEEauthorblockN{ Xinyue Peng\IEEEauthorrefmark{1}, Yanming Liu\IEEEauthorrefmark{2}, Yihan Cang\IEEEauthorrefmark{1}, Chaoqun Cao\IEEEauthorrefmark{1},  Ming Chen\IEEEauthorrefmark{1}\IEEEauthorrefmark{4}}
		\IEEEauthorblockA{\textit{*National Mobile Communications Research Laboratory, Southeast University, Nanjing, China} \\
  \textit{\IEEEauthorrefmark{2}ZheJiang University, Hangzhou, China} \\
		\textit{\IEEEauthorrefmark{4}Pervasive Communication Research Center, Purple Mountain Laboratories, Nanjing, China}\\
		Email:$\left \{\text{xinyuepeng, yhcang, chqcao, chenming}\right \}$@seu.edu.cn, $\left \{\text{oceann24}\right \}$@zju.edu.cn}
	}

\maketitle

\begin{abstract}
Solving non-convex resource allocation problems poses significant challenges in wireless communication systems, often beyond the capability of traditional optimization techniques. To address this issue, we propose LLM-OptiRA, the first framework that leverages large language models (LLMs) to automatically detect and transform non-convex components into solvable forms, enabling fully automated resolution of non-convex resource allocation problems in wireless communication systems. LLM-OptiRA not only simplifies problem-solving by reducing reliance on expert knowledge, but also integrates error correction and feasibility validation mechanisms to ensure robustness. Experimental results show that LLM-OptiRA achieves an execution rate of 96\% and a success rate of 80\% on GPT-4, significantly outperforming baseline approaches in complex optimization tasks across diverse scenarios.
\end{abstract}


\section{Introduction}
In wireless communication systems, resource allocation, which includes spectrum allocation, power control, and interference management, is crucial and widely applicable \cite{zhou2023survey}. Convex optimization offers stable and efficient solutions, especially when closed-form solutions are available. For instance, fractional programming effectively optimizes metrics like signal-to-interference-plus-noise ratio (SINR) and energy efficiency in applications such as wireless power control and beamforming \cite{b3}. However, many real-world problems are non-convex. In multi-cell interference scenarios, for example, non-convexity arises from coupled variables like transmit power, bandwidth, and discrete user scheduling, complicating optimal resource allocation. Addressing these issues often requires complex modeling, transformation, and relaxation techniques to convert non-convex problems into convex ones \cite{zhou2024large}, typically depending on expert knowledge  such as advanced mathematical optimization, signal processing principles, and wireless network design strategies.

To reduce reliance on expert knowledge, recent research shows that large language models (LLMs) show significant potential in tackling complex mathematical problems. For instance, LLMs achieve 97.1\% accuracy in solving mathematical challenges with zero-shot prompting on the GSM8K dataset \cite{zhong2024achieving}. They can also address classical optimization problems, such as linear regression and the traveling salesman problem, through iterative methods \cite{yang2024large}. Moreover, LLMs can leverage existing solvers like \texttt{Gurobi} and \texttt{CVXPY} for optimization tasks \cite{b2}. For example, the \textit{gurobi.abs} function enables direct modeling of $\ell_1$-norm objectives, allowing LLMs to avoid redundant auxiliary constraints and variables, thus accelerating the solving process \cite{ahmaditeshnizi2023optimus}. However, these solvers primarily handle convex problems, limiting their ability to solve non-convex challenges directly,  so LLMs must draw on their own capabilities to transform and solve non-convex challenges.

To enable more effective handling of complex optimization problems,  prompt engineering techniques\cite{rubin2022learning} are introduced to enhance their reasoning capabilities of LLMs. By adopting structured prompts, LLMs can engage in multi-step optimization processes, where techniques like chain-of-thought and tree-of-thought significantly improve their performance in solving complex mathematical issues \cite{he2023solving}.  Additionally, few-shot or zero-shot learning methods\cite{brown2020language} maximize the application effectiveness of LLMs under limited computational resources\cite{kojima2022large}.

Recent studies show that LLMs effectively address optimization complexities. The authors in \cite{chen2024diagnosing} demonstrate that LLMs can identify infeasible constraints and suggest methods to relax or remove them, streamlining the solution process. Additionally, LLMs have been applied in \cite{ahmaditeshnizi2023optimus} for modeling, code generation, and solving convex optimization problems. However, while these studies have predominantly focused on convex problems, the authors in \cite{lee2024llm} highlight LLMs' potential in non-convex optimization within wireless communications. But their approach is limited to a simple communication scenario and teaches LLMs only the numerical input-output relationships instead of solving the underlying mathematical problems,  which undermines generalizability to more complex non-convex problems.

\begin{figure*}[htbp]
    \centering
    \includegraphics[width=\textwidth]{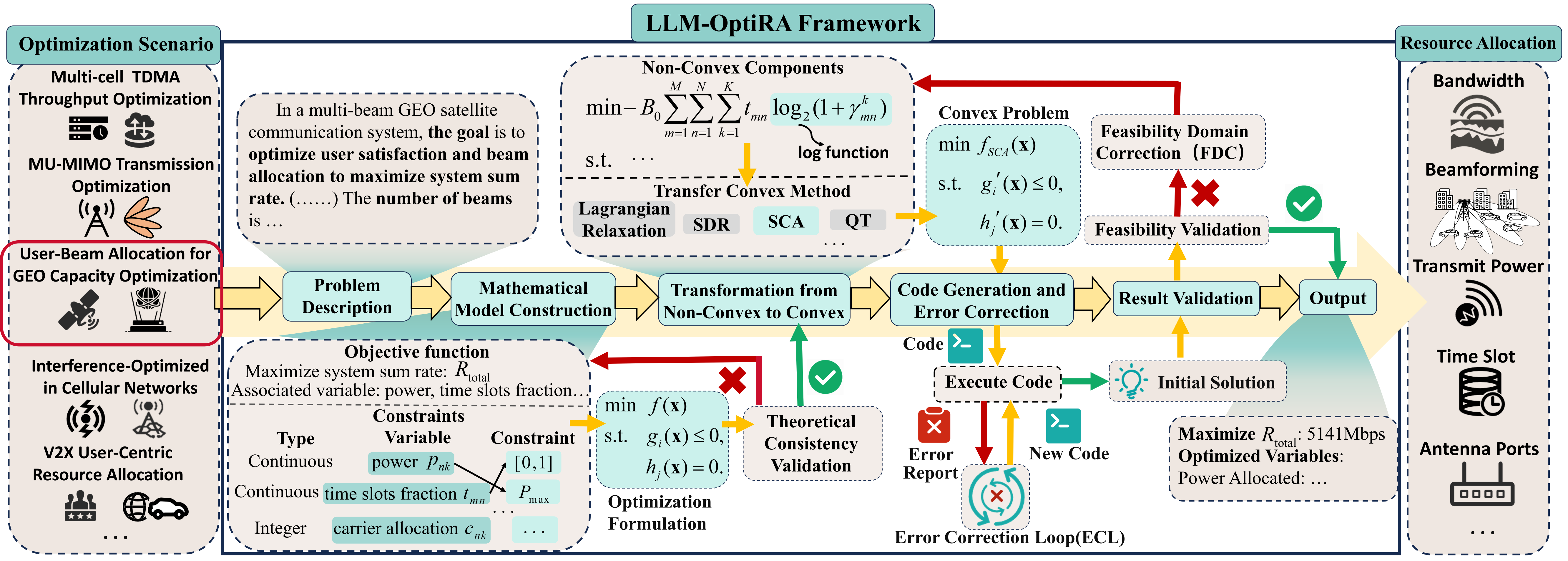}
    \caption{Overview of the LLM-OptiRA framework,   which is designed to optimize various scenarios for wireless resource allocation. It demonstrates the framework's key steps through the example of user-beam allocation in a GEO satellite system.}
    \label{fig:main}
\end{figure*}

To address the existing gap in applying LLMs to non-convex optimization problems in wireless communication systems, we introduce the LLM-OptiRA framework, which leverages advancements in mathematical reasoning, prompt engineering, and few-shot learning. The main contributions of this paper are summarized as follows.
\begin{itemize}
    \item We propose the LLM-OptiRA framework, the first application to LLMs to address the non-convex resource allocation problems in wireless communications. The framework automatically detects non-convex components and efficiently transforms them into convex problems, significantly reducing reliance on expert knowledge and enabling automated non-convex problem solving.
    \item We introduce Error Correction Loops (ECL) and Feasibility Domain Correction (FDC) as refine stage within the LLM-OptiRA framework. These enhancements not only boost the reliability and performance of LLM-OptiRA's solutions but also improve other existing advanced methods, demonstrating the framework's robustness in optimizing resource allocation problems.
    \item Experimental results show that LLM-OptiRA achieves an execution rate of {96\%} and a success rate of {80\%} on the GPT-4 model, significantly outperforming baseline methods. These results highlight LLM-OptiRA's effectiveness and robustness in solving non-convex problems in wireless communication systems.
\end{itemize}
Our implementation is openly available for research here: \url{https://github.com/Tibbers0310/LLM-OptiRA}

\section{LLM-OptiRA Framework}
This section details the specific steps of the LLM-OptiRA framework, as illustrated in Fig. \ref{fig:main}, with explanations based on the GEO satellite user-beam allocation example\cite{b17}.

\subsection{Problem Description}
In this step, users need to provide the LLM with a ``problem description" that includes the communication system background, optimization objectives, and the resource allocation requirements. Given that the LLM has foundational communication knowledge, users can present the problem in natural language without needing detailed formulas or specialized modeling languages.

\subsection{Mathematical Model Construction}\label{math} 
After the user inputs the problem, the LLM first analyzes the text and applies Named Entity Recognition (NER) methods\cite{wang2023gpt} to identify entities related to optimization. Through this approach, the model extracts key optimization variables \(\mathbf{x} = \{x_1, x_2, \dots, x_n\}\), such as power, time slots, and carriers.

Building on the extraction of optimization variables, the model identifies the objective function by analyzing the text for sentences that express the optimization goal. For example, if the goal is to maximize sum rate \(R_\text{total}\), the model locates relevant terms and identifies associated variables, such as transmit power \(x_1\) and time slots \(x_2\). This allows the model to create a set of objectives \(\mathcal{O} = \{R_\text{total}, x_1, x_2, \ldots\}\). With the objective set \(\mathcal{O}\) defined, the model synthesizes this information to derive the mathematical expression for the objective function \(f(\mathbf{x})\).

Next, the modeling of the constraints is conducted, for which the model needs to recognize the types of variables and the constraints. First, the model identifies the type of each variable \(t_i\) (e,g. integer or continuous), forming a set of extracted optimization variables \(\mathcal{E} = \{(x_i, t_i)\}\). Next, the model identifies the numerical values \(c_j\)(where \( c_j \in C \), \( C \) is the set of constraint values)  related to the constraints and then analyzes the text to determine which optimization variable \(x_i\) corresponds to each \(c_j\). Following this, the model extracts the sentences \(r_{ij}\) that mention these constraint relationships. Finally, these correspondences are placed into the set of constraint relationships \(\mathcal{R}_c\), defined as
\begin{equation}
    \mathcal{R}_c = \bigcup\limits_{x_i \in \mathcal{E},\, c_j \in C} \{(x_i, c_j, r_{ij})\}.
\end{equation}

Once the sets \(\mathcal{E}\) and \(\mathcal{R}_c\) are established, the LLM utilizes its expertise in communication theory and mathematical modeling to convert the elements of these sets into specific mathematical expressions, with the inequality constraints denoted as \(g(\mathbf{x})\) and the equality constraints represented as \(h(\mathbf{x})\).

Finally, to unify the representation of the optimization problem into a standard form, the optimization problem \(\mathcal{P}\) can be expressed as
\begin{equation}
\begin{aligned}
\mathcal{P:}\quad  \text{minimize } &f(\mathbf{x}) \\
    \text{s.t.}\quad
    &g_i(\mathbf{x}) \leq 0, \quad i = 1, 2, \ldots, m, \\
    &h_j(\mathbf{x}) = 0, \quad j = 1, 2, \ldots, n,
\end{aligned}
\end{equation}
where \(m\) represents the total number of inequality constraints, while \(n\) denotes the total number of equality constraints. 

\subsection{Transformation from Non-Convex to Convex}\label{convex}
In this step, the model first identifies non-convex components from the objective function and constraints. This process involves curvature analysis and assess the second derivative (or Hessian matrix). If the entire optimization problem is already convex, the model can proceed directly to the next solving step.

For identified non-convex components, the model  selects suitable convex transformation methods for approximation or relaxation, typically employing methods such as Semidefinite Relaxation (SDR), Successive Convex Approximation (SCA), or Lagrangian Relaxation \cite{b3}.
For example, in  maximizing 
 system sum rate problem, the coupling of SINR with logarithmic expressions results in non-convexity. This can be addressed using SCA, which employs a first-order Taylor expansion, allowing for the construction of a local convex surrogate around the current solution \(\mathbf{x}_{m}\). Consequently, the convexified objective function \(f_{\text{convex}}(\mathbf{x})\) can be expressed as
\begin{equation}
    f_\text{convex}(\mathbf{x}) = f_{SCA}(\mathbf{x}) = f(\mathbf{x}_{m}) + \nabla f(\mathbf{x}_{m})^{T} (\mathbf{x} - \mathbf{x}_{m}),
\end{equation}
where \(f_{SCA}(\mathbf{x})\) represents the sequential convex approximation, \(\mathbf{x}_{m}\) is the current solution, \(f(\mathbf{x}_{m})\) is the the objective function value at that point, and \(\nabla f(\mathbf{x}_{m})\) is the gradient. 

Constraints \(g_i(\mathbf{x})\) and \(h_j(\mathbf{x})\) are adjusted according to changes in the objective function. If they contain other non-convex features, they can be convexified utilizing the same principles. Ultimately, a complete convex optimization problem is obtained, allowing for the next steps in the solution process.

\subsection{Code Generation and Error Correction}
With respect to the convex optimization problems, the LLM generates corresponding Python code, denoted as \(C_{gen}\), to implement the optimization model. This code utilizes optimization solvers, denoted as \(\mathcal{S}\), to efficiently solve the optimization problem. While both solvers can handle convex problems, \texttt{CVXPY}\cite{b2} is typically used for standard convex optimization, whereas \texttt{Gurobi} excels in handling large-scale and complex linear programming and mixed-integer problems.

Next, the code is executed, and to check execution success, the metric \( \mathcal{Q} \) is defined: if the code runs without errors, \( \mathcal{Q} = 1 \); if errors occur, \( \mathcal{Q} = 0 \). Then if \( \mathcal{Q} = 1 \), the solution \( \mathbf{y}^* \) is obtained by minimizing the convexified objective function using the code \( C_{gen} \) and the solver \( \mathcal{S} \), expressed as \( \mathbf{y}^*=\min_{\mathbf{x}} f_{\text{convex}}(\mathbf{x}^*; C_{gen}, \mathcal{S}) \), where \( \mathbf{x}^* \) is the optimized variable. If the code execution fails, an error report \( \mathcal{R} \) is generated by the program, triggering the {Error Correction Loop (ECL)}, a component of the LLM-OptiRA framework designed to resolve errors.

The principle of ECL is to systematically analyze and adjust the reasons for code execution failure. In the \( k \)-th iteration, ECL examines the error report \( \mathcal{R} \)  to identify specific types of errors, such as incorrect parameter settings, mismatched constraints, or logical errors. Based on these findings, ECL then references the previously used solver \( \mathcal{S} \) to modify \( C_{gen} \), ensuring it meets the required conditions for successful execution. The updated code is re-executed until it runs successfully, at which point the new solution \( \mathbf{y}^* \) is obtained, or the maximum iteration count \( K \) is reached. This entire process, is represented as \( \mathcal{F}_{ECL}(\mathcal{R}, C_{gen}, \mathcal{S}, k) \).

 Consequently, the solution \( \mathbf{y}^* \) is expressed as
\begin{equation}
    \mathbf{y}^* = 
\begin{cases}
    \min_{\mathbf{x}} f_{\text{convex}}(\mathbf{x}^*; C_{gen}, \mathcal{S}) & \text{if } \mathcal{Q} = 1, \\ 
    \mathcal{F}_{ECL}(\mathcal{R}, C_{gen}, \mathcal{S}, k) & \text{if } \mathcal{Q} = 0 \text{ and } k < K,
\end{cases}
\end{equation}

\subsection{Result Validation}
Based on the solution of the optimization problem, verification is crucial for ensuring accuracy and effectiveness, which can be divided into two parts. 
\subsubsection{\textbf{Theoretical Consistency Validation}}
This validation
is set after the step of Mathematical Model Construction in \ref{math}. We utilize different criteria to evaluate various aspects of the generated formulas. The first criterion, denoted as \(\xi_1\), checks if the constructed formulas align with the problem description, confirming that the objective function is appropriately set for maximization or minimization. The second criterion, \(\xi_2\), ensures that no key variables or constraints are omitted. The third criterion, \(\xi_3\), verifies that the types of optimization variables match correctly. Finally, the criterion \(\xi_4\) assesses the accuracy of numerical values, including necessary measurement unit conversions.

LLM analyzes the problem context and the mathematical formulas to evaluate the four criteria, which can be collectively represented as the theoretical consistency metric \(\mathcal{T}\), expressed as
\begin{equation}
\mathcal{T} = \mathbb{H}(\xi_1) \cdot \mathbb{H}(\xi_2) \cdot \mathbb{H}(\xi_3) \cdot \mathbb{H}(\xi_4),
\end{equation}
where each \(\mathbb{H}(\cdot)\) is an indicator function that takes the value 1 if the corresponding condition is met, and 0 otherwise.

\subsubsection{\textbf{Feasibility Validation}}    
This validation is conducted  after successfully running the code and obtaining the solution \( \mathbf{y}^{*} \). While transforming a non-convex problem into a convex one, the model may relax constraints, potentially resulting in a solution that does not satisfy the original constraints. To ensure feasibility, the model substitutes the optimized variable \( \mathbf{x}^{*} \) into the original constraints, verifying \( g_i(\mathbf{x}^{*}) \leq 0 \) and \( h_j(\mathbf{x}^{*}) = 0 \). This process enhances the reliability of optimization results in practical applications, improving the utility of the LLM-OptiRA framework.

Consequently, the feasibility validation metric \( \mathcal{V} \) can be expressed as follows
\begin{equation}
    \mathcal{V} = {\mathcal{F}}_{val}(\mathbf{y}^{*} \mid g_i(\mathbf{x}^{*}) \leq 0, \ h_j(\mathbf{x}^{*}) = 0),
\end{equation}
where \( {\mathcal{F}}_{val} \) is the evaluation function that returns 1 if the solution meets the original constraints and 0 otherwise.

\subsection{Feasibility Domain Correction}
If the optimized variable \( \mathbf{x}^{*} \) is not within the feasible domain, the Feasibility Domain Correction (FDC) program is executed within the LLM-OptiRA framework.  The process is divided into two stages based on the maximum iteration count \( L \): in the first stage, the initial values of the optimization variables are adjusted; if no feasible solution is found after \( \left\lfloor \frac{L}{2} \right\rfloor \) iterations, the second stage begins, reanalyzing the original optimization problem.

In the first stage, the FDC program gradually adjusts the initial values \( \mathbf{x}_0 \) using the correction function \( \mathcal{F}_{adj}(\mathbf{x}_0, \Delta \mathbf{x}, \gamma) \). The adjustment \( \Delta \mathbf{x} \) and  \( \gamma \), representing
the step size for each adjustment, ensure gradual convergence to a feasible solution. After each iteration, the current solution's feasibility is checked using the metric \(\mathcal{V}\). If feasible (i.e., \(\mathcal{V} = 1\)), the process stops, yielding the final solution \( \mathbf{y}^{(final)} \).

If no feasible solution is found after \( \left\lfloor \frac{L}{2} \right\rfloor \) iterations, the model transits to the second stage, utilizing the reanalysis function \(\mathcal{F}_{re}(\mathcal{P}, l)\). Specifically, the model returns to the convexification process outlined in \ref{convex}, where it analyzes the original problem \(\mathcal{P}\), applies alternative methods for re-convexification. It then generates new solving code and executes it to obtain a new solution. If this solution is still infeasible, the process continues iterating until it either reaches the maximum count \(L\) or identifies a feasible solution, which then becomes the final solution \( \mathbf{y}^{(final)} \).

The entire process can be summarized by the following equations
\begin{equation}
    \begin{aligned}
    & \mathbf{y}^{(FDC)}_l = 
    \begin{cases} 
    \mathcal{F}_{adj}(\mathbf{x}_0, \Delta \mathbf{x}, \gamma) & \text{if } \mathcal{V} = 0 \text{ and } l < \left\lfloor \frac{L}{2} \right\rfloor, \\ 
    \mathcal{F}_{re}(\mathcal{P}, l) & \text{if } \mathcal{V} = 0 \text{ and } \left\lfloor \frac{L}{2} \right\rfloor \leq l \leq L, 
    \end{cases} \\ 
    & \mathbf{y}^{(final)} = \mathbf{y}^{(FDC)}_l \text{ if } \mathcal{V} = 1.
\end{aligned}
\end{equation}

Through this adaptive correction mechanism, the LLM-OptiRA framework effectively resolves complex optimization problems.

\begin{figure*}[t]
    \centering
    \includegraphics[height = 4.5cm]{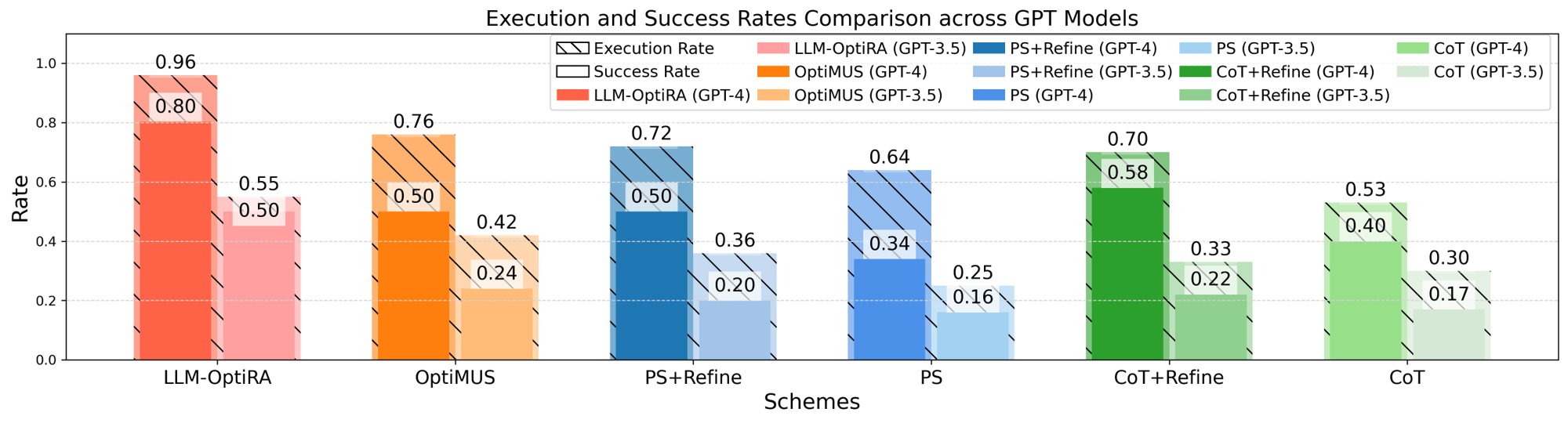}
    \caption{Comparison of execution and success rates for different schemes using GPT-4 and GPT-3.5 models.}
    \label{fig:exp1}
\end{figure*}

\section{Experiment and Analysis}
\subsection{Dataset}
To evaluate the performance of LLMs in solving non-convex optimization problems in wireless communication resource allocation, we need a corresponding dataset for training. Since no existing dataset addresses this issue in this field, we create the WireOpt dataset, which includes 100 optimization problems across various wireless communication scenarios. This dataset serves as the experimental input for assessing the performance of the LLM-OptiRA framework.


The dataset comprises 50 resource allocation problems extracted from textbooks, research papers, and Matlab documentation\cite{b3}\cite{Hou_Shi_Sherali_2014}\cite{MATLAB}, covering communication backgrounds, optimization formulations, and simulation parameters. To expand the dataset, the LLM learns core patterns and generates 50 additional problems in diverse contexts.

\subsection{Baseline Methods}
To evaluate our method holistically, we compare LLM-OptiRA with the baseline methods as

\textbf{OptiMUS}\cite{ahmaditeshnizi2023optimus}, a framework utilizing LLMs to formulate and solve Mixed Integer Linear Programming (MILP) problems from natural language descriptions. 

\textbf{Chain-of-Thought (CoT)}\cite{wei2022chain}, generates a step-by-step reasoning process,  which encourages the model to break down complex problems into smaller steps.

\textbf{Plan-and-Solve (PS)}\cite{wang2023plan}, enhances LLMs' reasoning by dividing problem-solving into  the model ``plans"  and  ``solves". 

\textbf{CoT+Refine}: Using the CoT method to solve optimization problems, along with the refine stage from LLM-OptiRA framework, which includes the ECL and FDC process.

\textbf{PS+Refine}:Using the PS method along with the refine stage from LLM-OptiRA framework.

\subsection{Evaluation Metrics}
The methods are evaluated based on two primary metrics: {Success Rate} and {Execution Rate}\cite{ahmaditeshnizi2023optimus}. For each optimization problem $\mathcal{P} \in \mathcal{D}$, a total of $N = 10$ evaluations are conducted using both models (GPT-4 and GPT-3.5), and the rates are computed as the average of these evaluations.

\begin{itemize}
    \item \textbf{Success Rate} is defined as the proportion of outputs that yield an optimal solution based on code execution and lie within the feasible domain while satisfying all constraints, expressed as 
    \begin{equation}
        \text{Success Rate} = \frac{1}{|\mathcal{D}|} \sum_{\mathcal{P} \in \mathcal{D}} \left( \frac{1}{N} \sum_{i=1}^{N} \mathcal{V}_\mathcal{P}\right),
    \end{equation}
    where \(\mathcal{V}_\mathcal{P} = 1\) indicates the solution of problem \(\mathcal{P}\) is feasible, and \(\mathcal{V}_\mathcal{P} = 0\) otherwise.

    \item \textbf{Execution Rate} measures the proportion of generated code that successfully executes and produces outputs, expressed as
    \begin{equation}
            \text{Execution Rate} = \frac{1}{|\mathcal{D}|} \sum_{\mathcal{P} \in \mathcal{D}} \left( \frac{1}{N} \sum_{i=1}^{N} \mathcal{Q}_\mathcal{P} \right),
    \end{equation}
   where \(\mathcal{Q}_\mathcal{P} = 1\) represents the code of problem \(\mathcal{P}\) is successful executed, and \(\mathcal{Q}_\mathcal{P} = 0\) otherwise.
\end{itemize}

\subsection{Performance Analysis of LLM-OptiRA}
Fig.\ref{fig:exp1} reveals the  key insights into the performance of the LLM-OptiRA framework, particularly in comparison to baseline methods, highlighting the execution and success rates across different configurations.

LLM-OptiRA significantly outperforms other baseline methods in both success and execution rates. It achieves an execution rate of 96\%, surpassing OptiMUS's 80\% in GPT-4, due to its robust code and effective error correction. Its success rate of 80\% in solving non-convex problems far exceeds OptiMUS's 50\%, thanks to its specialized non-convex to convex transformation module. Additionally, it outperforms CoT and PS schemes, with a success rate exceeding PS's 34\% and CoT's 40\%, highlighting its robustness and adaptability in complex optimization tasks.

Even on the GPT-3.5 model, LLM-OptiRA demonstrates impressive stability. The execution rate of LLM-OptiRA is {55\%}, surpassing OptiMUS's {42\%} and significantly outpacing CoT+Refine and PS+Refine, which stand at {33\%} and {36\%}, respectively. Although its success rate drops to {50\%}, it still outperforms OptiMUS at {24\%}, as well as CoT+Refine and PS+Refine, which achieve only {22\%} and {20\%}. This performance indicates that LLM-OptiRA remains effective even in models with lower capabilities.

The Refine mechanism, powered by the robust optimization capabilities of the LLM - OptiRA framework, significantly boosts the performance of CoT and PS. For instance, the execution rate of CoT rises from 53\% to 70\%, a remarkable increase. Alongside this, the success rate also experiences a significant improvement, clearly demonstrating that the structural advantages of our framework can effectively enhance the overall performance of other models in multiple aspects.

\begin{figure}[h]
    \centering
    \includegraphics[width=0.5\textwidth]{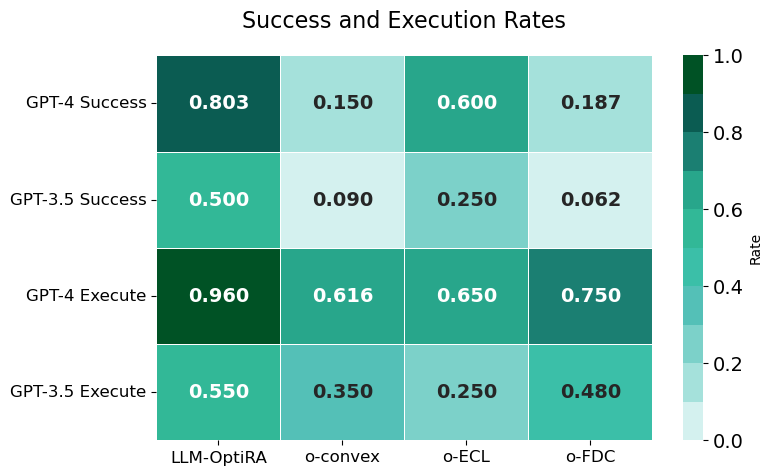}
    \caption{Impact of omitting key components of the LLM-OptiRA framework on success and execution rates across GPT-4 and GPT-3.5}
    \label{fig:heat}
\end{figure}

\subsection{Analysis of Ablation Experiments on LLM-OptiRA Framework Components}
The LLM-OptiRA framework consists of key processes, evaluating the impact of removing components on model performance. We now examine how these modifications compare to the complete framework and affect outcomes. The o-convex approach removes the convexification process. The o-ECL approach omits ECL process, while o-FDC does not include FDC process.

As shown in the Fig.\ref{fig:heat}, {o-convex significantly reduces both the success and execution rates, highlighting the importance of convexification within the LLM-OptiRA framework}.  Specifically, GPT-4's success rate drops from 0.803 to 0.15, reflecting a 79.8\% decline, while the execution rate decreases from 0.96 to 0.616. This highlights the critical need for convexification to transform non-convex problems into solvable forms.

{o-ECL negatively impacts the execution rate, which subsequently lowers the success rate.} For GPT-4, the execution rate drops from 0.96 to 0.65, a 32.3\% decrease, and the success rate declines from 0.803 to 0.6, reflecting a 25.3\% reduction. This decrease in execution rates leads to an increase in failed attempts to generate correct solutions, ultimately damaging the overall success rate.

{o-FDC primarily decreases the success rate, emphasizing its role in ensuring solution feasibility.} Experiments show that without FDC, GPT-4's success rate drops to 0.187, a 76.7\% reduction, while GPT-3.5's falls to 0.062, an 87.6\% decrease. Although GPT-4's execution rate remains at 0.75, the lack of FDC means that many solutions may not be within feasible regions, rendering them ineffective.

In summary, the absence of each component severely impacts the LLM's performance, demonstrating the importance of every step in the LLM-OptiRA framework for improving success and execution rates.

\subsection{Maximum Iteration Count Analysis of ECL and FDC}
\begin{figure}[t]
    \centering
    \hspace{-0.5cm}
\includegraphics[width=0.5\textwidth]{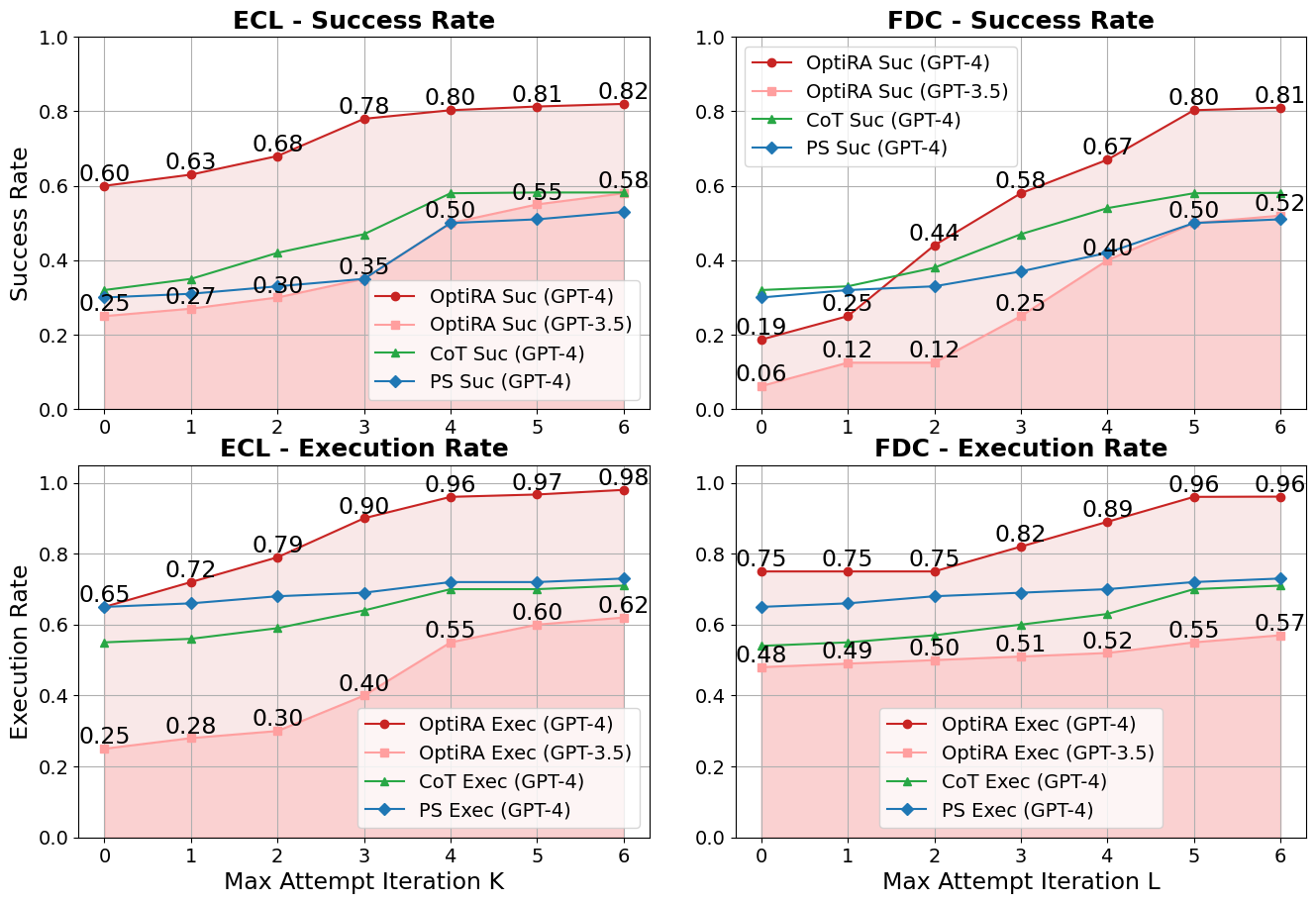}
    \caption{Success and execution rates for LLM-OptiRA under different maximum iteration counts for ECL and FDC.}
    \label{fig:iteration}
\end{figure}

The previous section highlights the significant roles that ECL and FDC play in the performance of the LLM-OptiRA framework. Therefore, this section delves deeper into their maximum iteration counts \(K\) and \(L\), respectively, examining their effects on the success and execution rates, as illustrated in Fig.\ref{fig:iteration}.

ECL significantly improves success and execution rates, especially in early iterations. For success rate, ECL rapidly identifies and corrects key errors, boosting GPT-4’s success rate from 60\% to 82\% and GPT-3.5’s from 25\% to 58\%, surpassing PS+Refine when \(K=4\), , indicating that ECL is effective for both high- and low-performance models. Execution gains are concentrated in initial iterations, with GPT-4 reaching 96\% by the fourth iteration. Despite GPT-3.5’s lower baseline, ECL raises its execution rate from 25\% to 62\%. To balance performance and computation, the framework sets \(K=4\).


FDC demonstrates notable improvements in success rate and maintains steady execution rate enhancements with increasing iterations. GPT-4’s success rate increases from 19\% to 81\%, and GPT-3.5’s from 6\% to 52\%, outperforming CoT and PS Refine. These gains stem from FDC’s iterative adjustment mechanism, which corrects solutions early and re-examines the problem at \(\left\lfloor\frac{L}{2}\right\rfloor\) iterations to fix infeasibility. Although FDC primarily focuses on ensuring solution feasibility, FDC also indirectly boosts execution rates by stabilizing the solving process, even under low initial performance. The framework sets \(L = 5\) to balance correction depth and efficiency.


\section*{Acknowledgment}
This work has been supported by the key research and development plan projects of Jiangsu Province under BE2022316 and Key R\&D Program of the National Ministry of Science and Technology (2023YFB2905603)

\section{Conclusion}
In this paper, we propose LLM-OptiRA, a groundbreaking framework that leverages LLMs to tackle non-convex resource allocation problems in wireless communication systems. Our experimental analysis shows that LLM-OptiRA achieves an execution rate of 96\% and a success rate of 80\% on the GPT-4 model, significantly outperforming baseline methods. The framework's success hinges on its integrated components as convexification, error correction, and feasibility checking, each of which plays a critical role in enhancing solution quality and overall robustness. Ultimately, LLM-OptiRA sets a new standard for automated resource allocation, demonstrating its capabilities in complex optimization scenarios.

	 \bibliographystyle{IEEEtran}
	 \small\bibliography{IEEEabrv,reference}

\end{document}